\documentclass[iicol]{sn-jnl}
\usepackage{times}
\usepackage{epsfig}
\usepackage{graphicx}
\usepackage{amsmath}
\usepackage{amssymb}
\usepackage{xspace}
\usepackage{xcolor}
\usepackage[justification=centering]{caption}
\usepackage{hyphenat}

\jyear{2024}%

\newcommand{\system}{Teledrive\xspace}
\begin{document}

\title{\system: An Embodied AI based Telepresence System}

\author{\centering\fnm{Snehasis} \sur{Banerjee}$^{1}$, 
\fnm{Sayan} \sur{Paul}$^1$,
\fnm{Ruddradev} \sur{Roychoudhury}$^1$,
\fnm{Abhijan} \sur{Bhattacharya}$^1$,
\fnm{Chayan} \sur{Sarkar}$^1$,
\fnm{Ashis} \sur{Sau}$^1$,
\fnm{Pradip} \sur{Pramanick}$^1$,
\fnm{Brojeshwar} \sur{Bhowmick}$^1$}
\email{snehasis.banerjee@tcs.com}
\email{p.sayan@tcs.com}
\email{ruddra.roychoudhury@tcs.com}
\email{abhijan.bhattacharyya@tcs.com}
\email{sarkar.chayan@tcs.com}
\email{ashis.sau@tcs.com}
\email{pradip.pramanick@tcs.com}
\email{b.bhowmick@tcs.com}

\affil{$^1$\orgdiv{Visual Computing and Embodied AI, TCS Research}, Kolkata, India}

\abstract{
This article presents `Teledrive', a telepresence robotic system with embodied AI features that empowers an operator to navigate the telerobot in any unknown remote place with minimal human intervention. We conceive Teledrive in the context of democratizing remote `care-giving' for elderly citizens as well as for isolated patients, affected by contagious diseases. In particular, this paper focuses on the problem of navigating to a rough target area (like `bedroom' or `kitchen') rather than pre-specified point destinations. This ushers in a unique `AreaGoal' based navigation feature, which has not been explored in depth in the contemporary solutions. Further, we describe an edge computing-based software system built on a WebRTC-based communication framework to realize the aforementioned scheme through an easy-to-use speech-based human-robot interaction. Moreover, to enhance the ease of operation for the remote caregiver, we incorporate a `person following' feature, whereby a robot follows a person on the move in its premises as directed by the operator. Moreover, the system presented is loosely coupled with specific robot hardware, unlike the existing solutions. We have evaluated the efficacy of the proposed system through baseline experiments, user study, and real-life deployment.}

\maketitle

\section{Introduction}

This article\footnote{Data Availability Statement:\\The data that support the findings of this study is available from \\ \url{https://github.com/facebookresearch/habitat-lab\#data}} primarily focuses on a telepresence robotic system in the context of remote caregiving. 
The pandemic situation demanded `social distancing' as the new normal. Yet careful monitoring of patients in isolation must be taken care of without risking the lives of `caregivers'. Even without the pandemic, there is a shortage of caregivers in different parts of the world, which is expected to be acute during the next pandemic outbreak~\cite{lal2022pandemic}. The availability of caregiver service must be done in a democratized manner such that individual care is possible for geographically distant individuals. A telepresence robot can address part of this issue. However, a major hindrance to the wider deployment of telepresence systems is the ease of use, particularly for a non-expert user. Existing telepresence systems, like Double3\footnote{https://www.doublerobotics.com/}, provide a manual navigation capability, which is often cumbersome for a user in a non-familiar or semi-familiar environment. Moreover, manual navigation requires continuous user intervention to move the remote robot from place to place within the remote environment. Furthermore, existing telepresence systems are tightly coupled with the robot hardware (like Amy\footnote{https://www.amyrobotics.com/}), which makes it difficult for enhancement, particularly by third-party developers. Thus, hardware-independent robot software development is needed for ease of feature addition, which has been exactly followed in this work through loosely coupled software modules.

\begin{figure}[t]
    \centering
    \includegraphics[width=\linewidth]{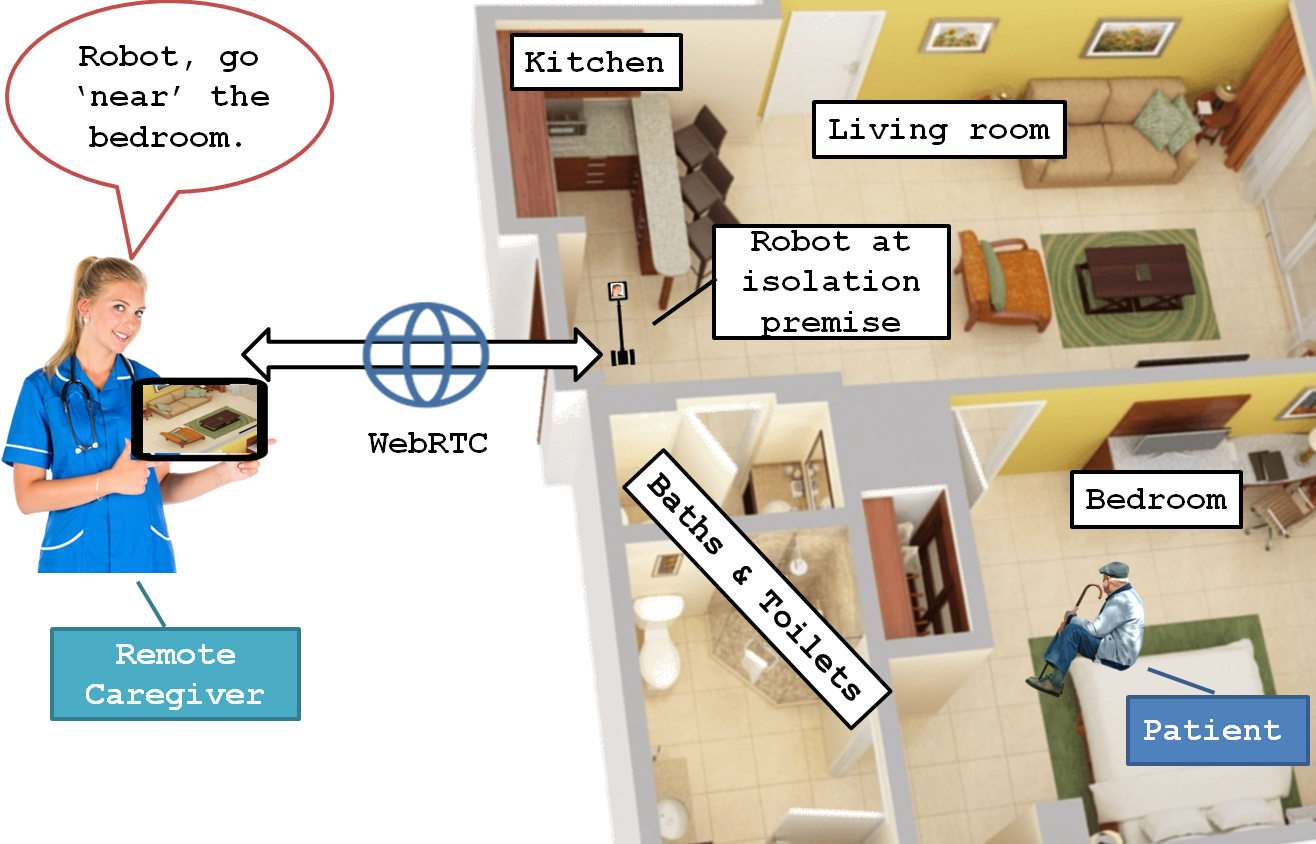}
    \caption{`Teledrive' system is given AreaGoal-based navigation instruction in remote caregiving use case.}
    \label{fig:exampleAreaGoalIntro}
\end{figure}

A telepresence system typically maintains a real-time connection with an application at the caregiver’s end and acts as an `avatar' of the caregiver at the patient's premise. The caregiver must navigate the patient’s premise in real-time through the Avatar interface based on audio-visual feedback as part of the ongoing real-time multimedia communication~(Fig.~\ref{fig:exampleAreaGoalIntro}). In most systems, the robot Avatar is maneuvered by the remote caregiver via manual instructions using on-screen navigation buttons, a joystick, a keyboard, etc. However, in an unknown premise (in the case of a tele-doctor), it would be too tedious for the caregiver to manually navigate the robot to the patient’s location. Hence, we conceive a system in which a caregiver can provide remote verbal instruction to the robot to navigate near a desired location inside the room (e.g., `bedroom'). The scope of the application can be further extended to isolation wards in health centers. Speech-based human-robot interaction (HRI) increases the usability and acceptability of the robot~\cite{pramanick2019enabling} in this context. Hence, a need is felt for a software architecture that can facilitate various embodied AI algorithms and machine learning models (and technology enablers) to work in harmony in the aforementioned scenario.

\subsection{Motivation}
Navigation is the key task of a telepresence robot. To establish the motivation for this work, let us take the help of the remote caregiver scenario depicted in Fig.~\ref{fig:exampleAreaGoalIntro}. Based on scene analysis on the ego view (front camera view), the robot is able to move to a position near the intended goal location. Once the robot reaches `near' that position, the caregiver can take manual control and perform finer control through the on-screen navigation buttons. In this example, the robot is connected with the caregiver’s PDA over the Internet through a WebRTC-based communication protocol. The robot is at the entrance of the patient’s premises. The old aged patient is in the bedroom whose vitals can be recorded by the robot like heart rate \cite{broj_hr2}, emotions \cite{brojo_hr1} etc. The caregiver verbally instructs the robot to navigate to the bedroom. Once the robot is able to locate itself around the bedroom, the caregiver can manually lead the robot to the bed where the old patient is waiting. This motivates us to develop a navigation capability for the following tasks - (a) PointGoal: going to a point location (b) ObjectGoal: searching and finding an object category (c) AreaGoal: navigating to a region category. Also, while navigating the environment, the robot may get confused about the next step or get stuck at a spot -- in such a scenario, it is essential to leverage dialogue exchange with the operator to clarify the robot's next move. Usually in a caregiver scenario or in a meeting, different users from varied remote locations need to converse and take control of the robot in turns. This requires a remote telecommunication protocol that can handle multiple users, also called multi-party communication. In this work, we have built a system catering to these requirements. To illustrate further, let us assume that a user gives a speech instruction `check if the cup is in the bedroom' which gets processed into a task instruction. In this case, the robot needs to invoke the AreaGoal task to find the bedroom region. Next, the robot needs to search for the object `cup' invoking the ObjectGoal task. It is to be noted that PointGoal is invoked in intermediate points for local transitions -- or intermediate points between the start and end goal. If there are two or more cups in the bedroom, the robot can ask a question back (dialogue exchange) to verify if indeed target goal is achieved.

Before deploying the algorithms in a real robot, the standard way is to train and test them on an actuation-enabled virtual robot in a simulator on some photo-realistic datasets of indoor scenes or in a created scene using SfM \cite{broj_sfm}. Matterport3D~\cite{chang2017matterport3d}, Replica~\cite{straub2019replica} and Gibson~\cite{xia2018gibson} are some popular 3D datasets mapped from real world house layouts. Among the choice of simulators for the tasks, we have used AI Habitat~\cite{savva2019habitat} (a simulation platform dedicated to Embodied AI research) as state-of-the-art benchmarks have also used the same framework. AI Habitat enables developers to test their algorithms in a 3D engine, where input is the robot's actuation commands and output is the actuation in the scenes loaded from the dataset. AI Habitat provides access to the ego view camera of the virtual robot in the form of an RGB frame, depth frame, and optionally semantic annotation of the scene. The odometry information is also available at each time step of the robot alongside the ego camera view.

The concept of Embodied AI is as follows - artificial agents or virtual robots, operating in 3D environments take as input egocentric perception and output actions based on a given goal task. This also includes training of embodied AI agents in realistic 3D simulators, before transferring the learned skills to reality. The presented telepresence system conforms to the embodied AI philosophy in the following manner. The user instruction is received using a human-robot interaction module where robot grounds the instruction \cite{broj_talk_vehicle} and ambiguity is resolved with respect to the observed scene using principles presented in~\cite{pramanick2022talk, pramanick2022doro}. The other three embodied AI downstream tasks, namely PointGoal, ObjectGoal, and AreaGoal, as defined in~\cite{anderson2018evaluation}, are taken care of -- initially in 3D simulators and then transferred to real-world deployment. Additionally, the person-following module is trained and tested in a 3D simulator before transferring to a real-world robot.

\subsection{Contributions}

The primary contributions of this paper are enlisted:
\begin{itemize}
\item We have concretely defined the problem specification of a set of `Areagoal' tasks and specified the metrics, benchmark, and approach with promising results. This is tied with individual contributions in Human-Robot Dialogue Exchange, Pointgoal, and Objectgoal tasks as enablers. To the best of our knowledge, this is the first work on the `AreaGoal' downstream task tested on benchmark, backed by a real-life deployment.
\item We have carried out a user study about the usability and functionality of the features presented in the Teledrive system and revealed interesting insights based on real human experience.
\item We have presented a software framework containing embodied AI features (including person following) that runs on ROS 1 or ROS 2 compliant robotic hardware in contrast to prior art where software features and hardware are tightly coupled~\cite{doi:10.1126/scirobotics.abm6074}.
\item We have developed a custom communication framework to enable real-time exchange of multimedia and control signals between the robot and the remote operator. This is achieved through custom exchange semantics built on standard WebRTC APIs. This enables an end-to-end web browser-based system. We have also built interfaces between the WebRTC APIs and the low-level robot control APIs. In lines of the operator-to-robot channel, we have created custom semantics on WebSocket for exchange between the Robot and the Edge in real-time for offloading of computation from the robot and fetching the computation outcome into the robot.
\item We have developed a device-agnostic responsive web browser-based frontend, that makes the solution platform-independent and easily accessible over URL by remote users having internet connectivity across mobile device variants.
\end{itemize} 

\begin{table*}
    \centering
    \caption{Comparison of features in existing telepresence systems.}
    \label{tab:comparison_variations}
    \rotatebox{90}{
    \begin{tabular}{|p{2cm}|l|l|p{1.1cm}|p{0.55cm}|l|p{0.5cm}|p{0.85cm}|l|p{1.5cm}|p{1.05cm}|l|p{0.8cm}|p{1cm}|l|}
    \hline
        Features & Ours & Double & ENRICH ME~\cite{cocsar2020enrichme} & Amy~\cite{amy} & Kubi~\cite{wu2017evaluation} & Ava~\cite{lewis2014evaluating} & PadBot~\cite{padbot} & Vgo~\cite{tsui2011exploring} & Ohmni~\cite{ohmni} & BotEyes~\cite{boteyes} & Giraff~\cite{orlandini2016excite} & Beam~\cite{beampro} & FURoI~\cite{lutz2016privacy} & Temi~\cite{hung2021foodtemi}\\ \hline
        \nohyphens{Platform independence} & y & n & n & n & n & n & n & n & n & n & n & n & n & n \\ \hline
        \nohyphens{Browser\;\;\;\;\;\;\; GUI} & y & y & n & n & n & n & n & y & y & y & n & n & n & n \\ \hline
        \nohyphens{Manual navigation} & y & y & y & y & n & y & y & y & y & y & y & y & y & n \\ \hline
        \nohyphens{AR goal based navigation} & y & y & n & y & n & n & n & n & n & n & n & n & n & y \\ \hline
        \nohyphens{Map based navigation} & y & n & y & n & n & y & n & n & n & n & n & n & n & y \\ \hline
        \nohyphens{Area goal based navigation} & y & n & n & n & n & n & n & n & n & n & n & n & n & n \\ \hline
        \nohyphens{Speech based navigation} & y & n & n & n & n & n & n & n & n & n & n & n & n & y \\ \hline
        \nohyphens{Information mashup} & y & n & n & n & n & n & n & n & n & n & n & n & n & n \\ \hline
        \nohyphens{Speaker localization} & y & n & n & n & y & n & n & n & n & n & n & n & n & n \\ \hline
        \nohyphens{Face identification} & y & n & y & y & n & n & y & n & n & n & n & n & y & y \\ \hline
        \nohyphens{Person following} & y & n & n & n & n & n & n & n & n & n & n & n & n & y \\ \hline
        \nohyphens{Multi-party federated control} & y & n & n & n & n & n & n & n & n & n & n & n & n & n \\ \hline
        \nohyphens{Automatic map generation} & y & n & y & n & n & n & n & n & n & n & n & n & n & y \\ \hline
        \nohyphens{Dialogue based disambiguation} & y & n & n & n & n & n & n & n & n & n & n & n & n & n \\ \hline
    \end{tabular}
    }
\end{table*}

The overall organization of the paper is as follows. Section~\ref{related_work} provides a comparison table of the state-of-the-art telepresence systems, highlighting relevant related works.  In section~\ref{secSysArch}, we present the software system architecture. Section~\ref{secCogDialogue} discusses task understanding and dialogue exchange with robots. Section~\ref{secCogNavigation} describes various aspects of cognitive navigation with a focus on AreaGoal and `person following' tasks. `PointGoal' and `ObjectGoal' tasks are also discussed in short. Section~\ref{secUI} describes the browser-based remote operator interface and user studies. Finally, we conclude the paper in section~\ref{secConclusion} along with future work.

\section{Related Work}
\label{related_work}

There has been a significant amount of work done in the robotic telepresence catering to a multitude of solutions~\cite{melendez2017web} \cite{tuli2020telepresence} \cite{soares2017mobile} \cite{herring2013telepresence} \cite{ng2015cloud} \cite{michaud2007telepresence} \cite{tan2019toward} \cite{cesta2016long} \cite{monroy2017integrating} \cite{beno2018work} \cite{kristoffersson2013review}. The majority of telepresence systems are intended for use in an office setting. Some are suited for elderly care, healthcare, and robotic research. Most of the prior art has focused on the hardware capabilities and allied features. However, we focus on the software platform itself that can run on any ROS-compliant hardware with some adaptation at the hardware level. In terms of software, we have included several components to perform embodied AI tasks to give a flavor of cognitive intelligence to the telepresence scenario. The closest work is ENRICHME~\cite{cocsar2020enrichme} which supports some simple software-level cognitive features. However, we offer a significant shift towards complex and sophisticated embodied AI features that have been tested to work on a real robot.

Standalone efforts on embodied AI tasks have recently gained the focus of the research community - like cognitive navigation~\cite{duan2022survey} and human-robot dialogue. A number of works on Speech based HRI in robots have focused on accessibility~\cite{tsui2015accessible} and cloud infrastructure for HRI using speech~\cite{ deuerlein2021human}, however, the area of speech HRI for general purpose telepresence scenario has limited peer-reviewed work. In cognitive navigation, for the `PointGoal' problem, there exists some work that uses frontier-based planners~\cite{batinovic2021multi, chattopadhyay2021robustnav}. There has been some work on the person following by robot~\cite{personFollow1, cheng2019person}, however, our work gives control to the remote user in order to follow a person, making it suitable for the telepresence scenario; and introduces engineering contributions to execute the task in the real world. In recent times, the `ObjectNav' task has grabbed the attention of the robotics research community. The initial work on Semantic Visual Navigation~\cite{yang2018visual} uses scene priors learnt from the standard image dataset. Their work uses Graph Convolutional Networks (GCNs) to embed prior knowledge into a Deep Reinforcement Learning framework and is based on an Actor-Critic model. However, that work lacked a concrete decision model when two or more objects are in the same scene. The proposed method makes an improvement on GCN-based ObjectNav~\cite{tatiya2021knowledge} by using trajectory data priors along with a learnt region-object joint embedding learned from a GCN training to perform the ObjectNav task. We have presented comparative results with the end-to-end RL-based ObjectNav -- SemExp~\cite{chaplot2020object} to support the efficiency of the approach. In the context of the `AreaGoal' problem, there is no dedicated work, apart from some work on scene graph analysis~\cite{liu2021toward} to decide the region. Another work on robot navigation learning is based on sub-routines from egocentric perception~\cite{kumar2020learning} that shows an example run for the AreaGoal task in the `washroom' region. However, the work neither gives away details of the AreaGoal task specifically nor performed any benchmark studies on the AreaGoal task. Their work was more focused on the PointGoal task. In contrast, our paper tackles the AreaGoal problem in great detail backed by results. Hence, the combination of multiple software features within the telepresence framework is a unique proposition of this work, which is further evident from the comparison of features in prior art as presented in table~\ref{tab:comparison_variations}.

\section{System architecture}
\label{secSysArch}
\begin{figure*}[h]
    \centering
    \includegraphics[width=0.8\linewidth]{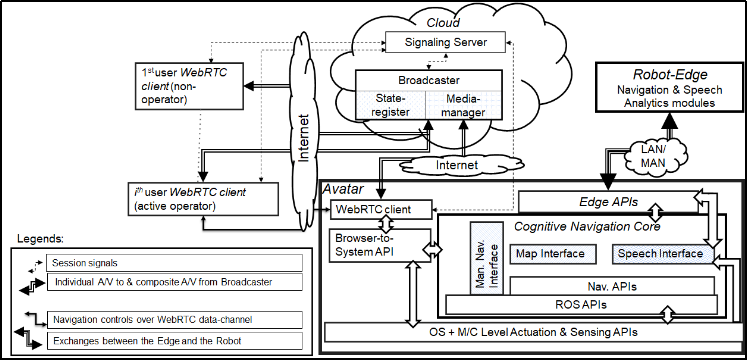}
    \caption{Edge-Cloud topology of the distributed networked Embodied AI of \system.}
    \label{fig:sysarchHigh}
\end{figure*}

\begin{figure*}[h]
    \centering
    \includegraphics[width=\linewidth]{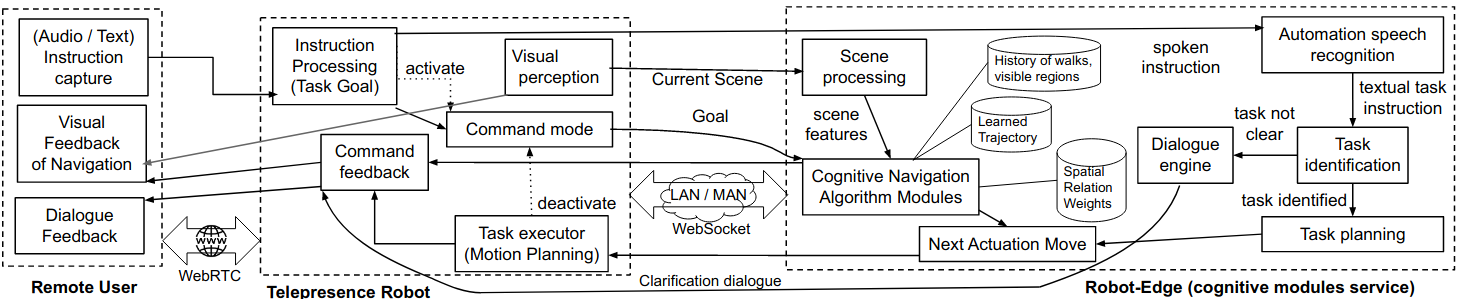}
    \caption{System architecture of \system showing distributed set of building blocks spread across four entities -- master device, cloud server, robot (avatar), and robot-edge.}
    \label{fig:archi_high_level}
\end{figure*}

In this section, we provide a high-level overview of \system (Fig.~\ref{fig:sysarchHigh} and Fig.~\ref{fig:archi_high_level}). The edge-cloud hybrid architecture is based on the principles laid out in~\cite{bhattacharyya2022teledrive, sau2021teledrive, sau2023edge}. It comprises four major subsystems -- Communication, Embodied Dialogue Exchange, Embodied Navigation, and User Interface. 

Robotic Telepresence enables virtual presence of a distant human (aka Operator) through an in-situ Robot (aka Avatar) which is maneuvered in real-time by the Operator, while having multimedia conferencing with the Avatar side participants over the Internet. The recent pandemic has underscored the importance of such systems. The Edge-enabled architecture helps offload computation required for the cognitive tasks by the robot. The Edge may be an in-situ computer connected over local WiFi or it may be inside the network provider’s infrastructure. The communication mechanism coupling the entire system has two major aspects: (1) to realize collaborative multi-presence session (CMS) and (2) to coordinate amongst computation modules, distributed between the Edge and the CMS on the Avatar. The CMS is maintained by WebRTC compatible browsers on every peer. The first part is ensured by a unique application layer on WebRTC. It supports a unique hybrid topology to address diverging Quality of Service (QoS) requirements of different types of data. The A/V is exchanged between the Avatar and the remote users through a cloud centric star topology over the SRTP based WebRTC media channel. But the delay-sensitive control signals from the active Operator to the Avatar is exchanged over a P2P data-channel on SCTP established directly between the Operator and Avatar on demand. This is unlike usual WebRTC topologies which support either mesh or star topologies using its inherent P2P (peer-to-peer) mechanism.

Navigation, though natural to humans is non-trivial for robots. The `PointGoal' problem is solved using the egocentric views obtained from the robot, which are then combined to make a global view of the environment using a spatial transform. The global top-view map is then used to plan a trajectory using the A* planner. The Dialogue engine helps in disambiguation of tasks instructions specified verbally using a pre-trained model ‘wav2vec’, which is passed on to task identifier~\cite{sarkar-2023-tage} to understand the goal and invoke the corresponding task module as a service.

In the context of the caregiver example, the telepresence system maintains a real-time connection with an application at the caregiver’s end and acts as an Avatar of the care giver at the patient’s premise. The caregiver must navigate the patient’s premise via robot avatar in real-time, based on audio-visual feedback. The caregiver can provide a remote verbal instruction to the robot to navigate near a desired location inside the room (e.g., `bedroom'). The speech based human-robot interface understands the desired goal from the voice. Based on the in-situ analysis of the semantic map derived from the live captured frames, the robot is able to move to a position near to the intended location. Once the robot reaches near that goal, the caregiver can take manual control and perform finer control through on-screen navigation buttons. The robot is connected with the caregiver’s mobile device over the Internet through the WebRTC based communication protocol. The following sections discuss the individual building blocks in detail, which are needed to support the caregiver telepresence scenario, amongst other possible use cases.

\section{Human Robot Interaction}
\label{secCogDialogue}
\begin{figure*}[h]
    \centering
    \includegraphics[width=0.95\linewidth]{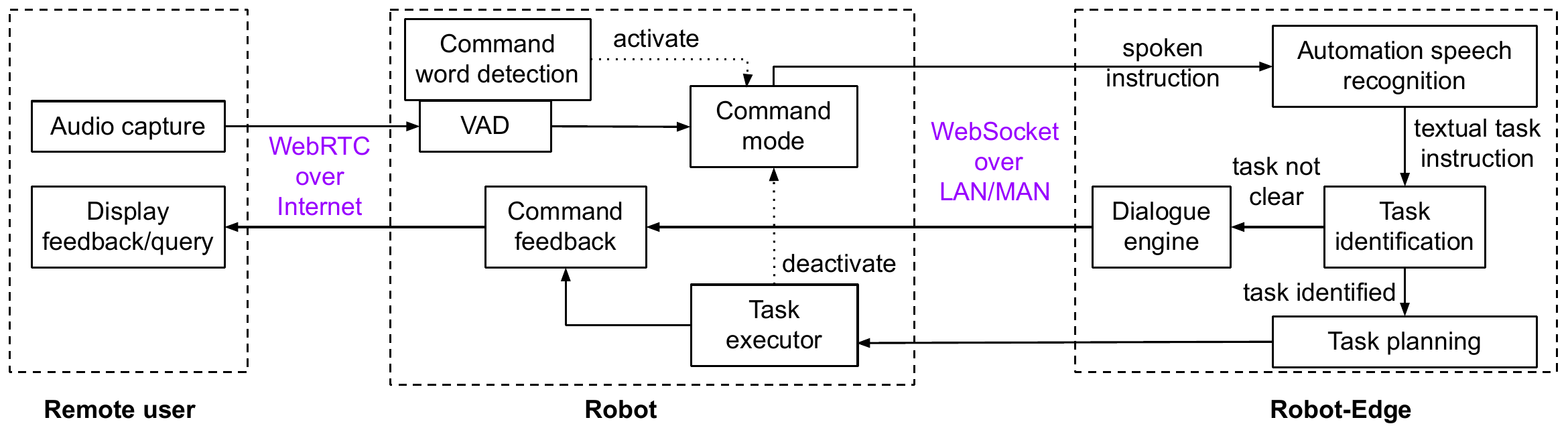}
    \caption{Embedding of human-robot interaction mechanism into \system.}
    \label{fig:archi_hri}
\end{figure*}
In a telepresence system, the robotic platform is primarily used to interact with human beings. Instead of manually controlling the robot through a continuous set of control commands, autonomous action by the robot based on high-level command is quite desirable~\cite{pramanick2019your, pramanick2019enabling}. This requires a robust human-robot interaction (HRI) module.

Fig.~\ref{fig:archi_hri} depicts the HRI module embedded with our Teledrive system. Since the speech interface is the most intuitive and user-friendly way to interact with a robot, Teledrive also supports speech-based task instruction to the robot. However, most of the time what the master says is meant for the remote audience. To distinguish the control instruction for the robot and ordinary speech, we devised a protocol of prefixing the control instruction with a command word. This is common for many personal assistant systems where a wake word is spoken first, before instructing the assistant. Now whenever the robot receives an audio signal, the voice activity detection (VAD) module processes to find the command word. If there is no command word, no further action is taken. On the other hand, if the command word is detected, the command mode is activated. The command word is detected using a CNN-based binary classifier. The model takes any audio chunk of a specified duration (upper limit to pronounce the command word) and outputs whether the audio chunk contains the command word or not.

In the command mode, the received audio is recorded locally on the robot until a silence of significant duration is detected by the VAD, which signifies the end of instruction. Then, the recorded audio is transmitted to the robot edge for further processing. At the robot edge, the audio is translated to text first. For this, we use our own automated speech recognition (ASR) model for embodied agent~\cite{pramanick2022visual, pramanick2023kb}. This is an adaptation of the `wav2vec' ASR model, which is trained using a transformer-based deep neural network~\cite{schneider2019wav2vec}. Then, the textual instruction is passed to a task identification module. This module identifies the task type and the parameter(s) from the natural language instruction. An example is `navigate to the bedroom' where the task is navigation and the parameter is `bedroom' mapped to an area lookup table. If the task understanding is successful, the execution plan is communicated to the robot, where the plan is executed through API calls. At this point, the robot exits from the command mode. In case the robot edge fails to decode the task instruction, it generates a query for the operator (Master) that is passed onto the operator's devices through the robot communication channel.

Another aspect is handling ambiguity in HRI. Following the principles set in the work~\cite{pramanick2022talk}, the ambiguity in goal of the task instruction can be categorized, and accordingly the type of question back can be formulated to aid the robotic agent in selecting the next course of action to complete the task.

\section{Embodied AI Navigation Tasks}
\label{secCogNavigation}
One of the principal tasks that an embodied AI agent needs to perform very well is navigation. In this regard,~\cite{anderson2018evaluation} has classified the navigation tasks into \textit{PointGoal} (go to a point in space), \textit{ObjectGoal} (go to a semantically distinct object instance), and \textit{AreaGoal} (go to a semantically distinct area). The \textit{AreaGoal} problem (also called \textit{AreaNav}) has been elaborated here, while other tasks are described in brief. We evaluate the `ObjectGoal' navigation task results on $4$ evaluation metrics as specified in~\cite{anderson2018evaluation} and~\cite{chaplot2020object} and refer the metrics in Appendix (section~\ref{appendix1}).

For experimentation and benchmark perspective, a fixed discrete action space is the norm followed by the peer community for embodied AI tasks. However, in real life deployment in robots, the step length and the rotation angle can be specified, provided training is done on a wide action space. For the person following task, the robot is free to move and rotate by values supported by the software running on robot hardware. So we could specify precise angle of rotation and step length of movement in that case. For a robot with a camera only in the front, a backward motion is not allowed to prevent (a) collision (b) uncertainty as very little information comes into view due to reverse motion. So alternatively, a backward motion is done as $180^0$ turn in left or right, and then going forward.

The baselines for Cognitive Navigation are: (a) Random exploration: the robot does random movements, until the goal is reached or there is a timeout. However, this includes collision avoidance intelligence. (b) Frontier-based exploration: The robot tries to explore the entire layout of an indoor house, and while doing so will encounter the target goal at some point of time. This in some way is a frontier search method, avoiding visiting the same place again, and the success mostly depends on how close the target area is to the randomly initialized robot pose. Usually an upper limit of number of steps is kept depending on the dataset indoor space layout, and target needs to be reached within those steps. `PointNav' task is leveraged by both `ObjectGoal' and `AreaGoal' modules in order to move from one point to another pre-specified point location (by performing local motion planning). To ensure that during navigation, the robot maintains the trajectory with best possible network connectivity, we have used a zero learning, source-agnostic approach as presented in~\cite{ganguly2024sensing}.

\subsection{PointNav Module}
\label{PointNav}

\begin{figure}[h]
    \centering
    \includegraphics[width=\linewidth]{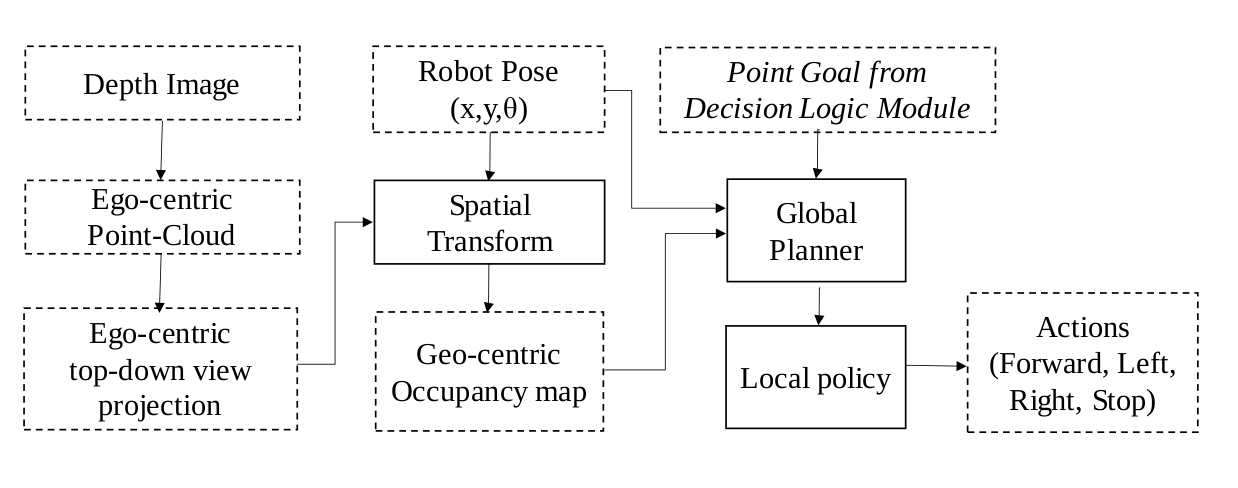}
    \caption{Method overview of PointNav Module.}
    \label{fig:pointNav}
\end{figure}

Given a start location ($x_1$,$y_1$) and an end goal location ($x_2$,$y_2$) for a ground robot, the task of this module is to help the robot navigate to the target within a certain predefined threshold without colliding with any static obstacles present in the environment. The robot is provided with an RGB-D camera sensor and wheel encoders that gives at each timestep -- the RGB image, depth image (raw or refined \cite{brojo_hd}) and absolute pose of the robot with respect to the start location. The depth image is converted to an egocentric point cloud using known camera intrinsics, and then is further flattened along the vertical axis (z-axis) to obtain an egocentric top-down projection occupancy map, containing the obstacles that robot has perceived from its ego view.

As this problem is solved using a differential drive robot executing a motion on flat surface, the absolute pose of the robot with respect to the start can be resolved into 3 variables: (x, y, o) -- x being the forward displacement, y being the displacement to the left perpendicular to x and o being the anti-clockwise rotation of the robot with respect to it's center. Given the wheel encoder differences at each timestep, the values of (x, y, o) can be defined using~\cite{odometry}. Our experiments with the Double3 Robot show, that on a flat surface (coherent with office, or home environments with hard carpet), the wheel odometry of Double3 has an average drift of 1 cm per metre of motion; and rotation error is negligible. In absence of a motion capture system or LiDAR, this was done by running the robot manually over loops of 3 metres to 10 metres so that that the start and end positions of the trajectory is the same; and then measuring the drift accumulated by the robot wheel odometer. We reason that since each subtask of the Area-goal problem (or any similar problem) is independent, and the absolute position of the robot is not needed across the whole trajectory, if we can accumulate the pose from trajectory's start position, till the trajectory ends, the purpose is served.

This egocentric map is in camera co-ordinate space, and is transformed to the world co-ordinate space using the absolute pose at that timestep by the spatial transform block (ST) to update the geocentric map. The Spatial transform block~\cite{chaplot2020learning} transform each point in the egocentric point cloud into it's position with respect to the start location. This updated geocentric map, current location and the goal location is fed to the Global Planner at every timestep t, which generates a feasible short but safe trajectory or path from the current location to the goal location. The global planner used here is a modified implementation of the A*  Algorithm~\cite{russell2016artificial}, where the free-space cost map is weighted more near the obstacles. As a result, the generated trajectories or paths are planned at a safe distance from the obstacles. It tries to find a good balance between finding the shortest path to the goal and avoiding collisions with obstacles. 
This planned path and the current pose is then fed to the Local Policy which is a heuristic planner that finds the next best action to be performed by the robot to align and follow the path generated by the global planner. The heuristic planner~\cite{chaplot2020learning} follows the global plan waypoints by deciding one of the actions out of `forward, turn left, turn right', by understanding it's orientation with respect to the nearest global planner waypoint.

We use a modular approach to solve this `Pointnav' task where mapper, global planner and local policy is stacked one after the other. This decoupled approach helps our method to generalize to different environments without the need for re-training or fine-tuning as compared to end-to-end learned approaches. It also helps in sim-2-real transfers. We use AI Habitat~\cite{savva2019habitat} as our evaluation framework. The habitat Pointgoal challenge 2020~\cite{habitat2020sim2real} uses a standard split for the Gibson Dataset. We have assumed ground truth pose for our evaluations. We use the validation split of the Gibson Dataset~\cite{xiazamirhe2018gibsonenv} which consists of 994 episodes where the robot is spawned at a random environment at a random start location and then given a random goal location that is reachable from the start point. An episode is deemed successful if the robot reaches the goal within 500 steps from the start location. It is assumed to have reached the goal if it gets within 0.36 metres (twice the robot radius which is 0.18 metres) of the goal location. We also tested this module on our real robot -- Double3~\cite{DoubleRobotics} which is equipped with Intel Realsense RGB-D cameras, however due to the lack of motion capture systems, we have not provided results to our evaluation on the real robot for this subtask here. However, we plan to provide an evaluation in future. We kept the above action space same in both simulation evaluation and real-world testing for smoother sim-2-real transfer. The results on AI Habitat with Gibson dataset are presented in the Table~\ref{table:results_pointnav}.

\begin{table}[t]
\caption{Results for `PointNav' task.}
\label{table:results_pointnav}
\centering
\begin{tabular}{p{0.3\linewidth}p{0.3\linewidth}p{0.3\linewidth}}
\hline
Number of Trials & Successful trials & Success (\%) $\uparrow$\\
\hline
994 & 905 & 91.05\\
\hline
\end{tabular}
\end{table}

\subsection{Visual Semantic Map Module}
\label{visSemMapsection}
\begin{figure}[h]
    \centering
    \includegraphics[width=\linewidth]{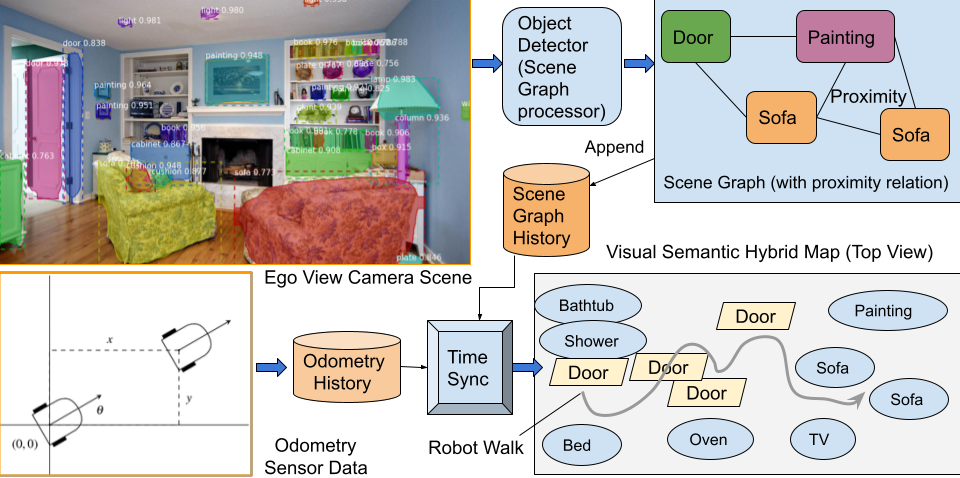}
    \caption{Semantic map generation from scene graph sequences learned from random walks in simulator.}
    \label{fig:sem_map}
\end{figure}
\begin{figure}[h]
    \centering
    \includegraphics[width=\linewidth]{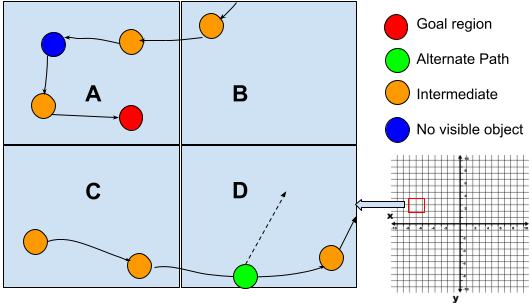}
    \caption{Grid Map for back tracking to the next unexplored landmark point for `AreaGoal' task.}
    \label{fig:grid_area_map}
\end{figure}

When the robot is moving in an unknown environment, it is imperative to create a map of its surroundings while it is moving. While a traditional Visual SLAM-based approach~\cite{taketomi2017visual} helps in `PointNav' problem, however, for tasks needing higher level semantic understanding of the scene, a metric map or voxel map is not sufficient. Hence, an alternative Visual Semantic Map is introduced that uses odometry data of the robot combined with perception (via RGB camera) to create a map that is both metric (having relative distance level granularity) and topological (retaining connection between scene graphs). The map is further enhanced with external semantic level features to link regions and objects in the robot's ego view. As seen in Fig.~\ref{fig:sem_map}, the history of scene graphs extracted from a sequence of scenes along with their time synced odometry data helps in generating the map incrementally. It is to be noted that raw image data from camera is processed to build the scene graph, and not semantic factual information as in the case of stream reasoning~\cite{mukherjee2013towards}~\cite{banerjee2018system}. Scene graph processor takes in the RGB camera image as input and applies object categorization using YOLO~\cite{jiao2019survey}. It populates a graph with nodes as objects and edges as proximity relation between objects. The initial position of the robot is taken as (0,0,0) in 3-D co-ordinate space facing east. In case the robot is navigating in the same floor or elevation, the z-axis (height) will be always positive, whereas translation of robot will happen in (x,y) plane. The `z' axis data points observed from ego view helps in aiding the robot to look in specific portions of 3D space while searching an object. As an example, `paintings' should be hanging on a wall at some height above floor, where as `sofa' will be grounded on the floor. 

The specific data-structure used for this is a rectangle grid as shown in Fig \ref{fig:grid_area_map}. It is computationally inefficient to search each and every point when the robot needs to look for the next unexplored landmark. Hence the map is initialized with a grid structure which is a large enough area in comparison to a regular house layout. Each grid cell has a set of Boolean tags -- explored or unexplored; having alternative branch path or not. Each grid cell can consist of multiple points from where the robot has taken an observation -- this also include the objects identified in the view. Each of the points has a view angle associated with it. In a navigation run of moving forward, the side view objects and openings which were kept out of view will be ignored. So later, if backtracking is needed, the information of those unexplored view angles can be used for further inspection in those areas. If a robot has reached a dead end, say $(x,y,\theta: 2,3,30^0)$, and need to backtrack, it can find the nearest cell to backtrack from the list of alternative path points (shown in green color in Fig. \ref{fig:grid_area_map}) kept in a list. Also, when the robot is exploring, it can look ahead a few grid cells (if previously visited) to find the best grid cell to move to maximize the objective of the specific long term navigation task. The green circles are alternate odometry points that the robot has visited in past. The orange points are the transition points in each step, assuming discrete steps of a fixed distance. Blue circle denotes a position, where the robot could not see any visible object in that view (may be free space like hall or a failure in object detection algorithm to identify any object). In such situation, it needs to take the next move randomly by invoking the PointNav module to come out of that situation of indecision. The red circle is the location where the robot needs to reach currently -- it can be the final goal or an intermediate goal given by the PointNav module to make it come out of a stuck region. Hence, this module helps in taking decisions where past navigation history and observations can be leveraged, by querying this representation.

\begin{figure*}[h]
    \centering
    \includegraphics[width=0.85\linewidth]{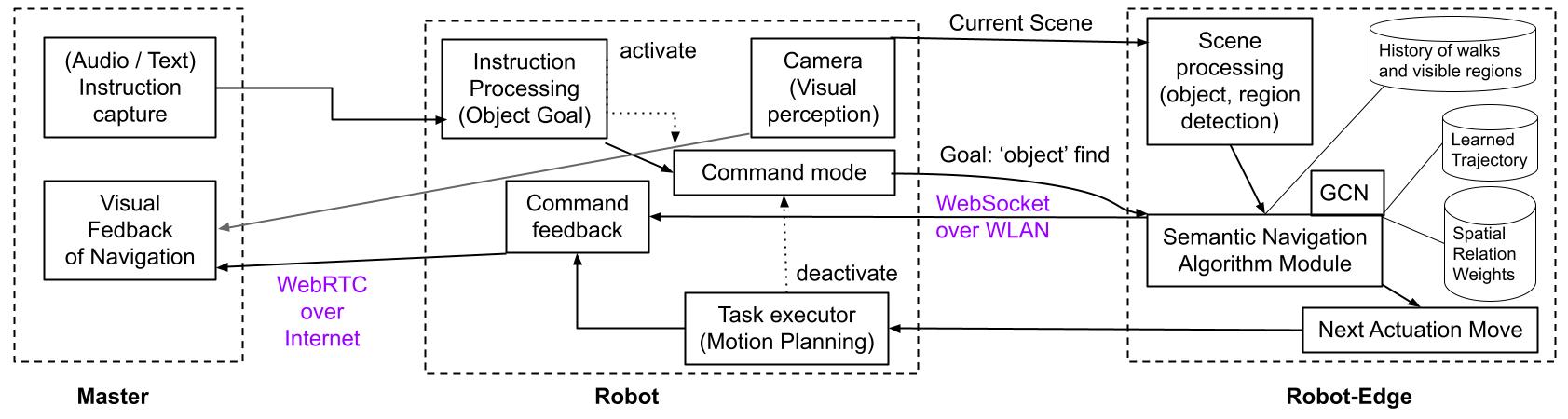}
    \caption{Semantic navigation for `ObjectGoal' task in telepresence scenario.}
    \label{fig:arch_semnav}
\end{figure*}

\begin{figure}[h]
\centering
  \includegraphics[width=0.8\linewidth]{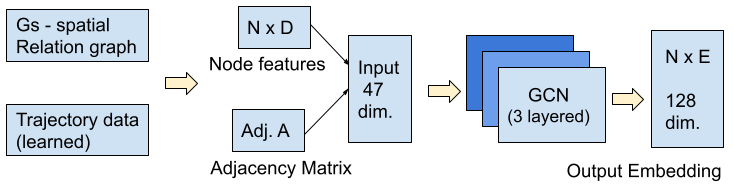}
  \caption{Training GCN to encode embedding of information in SRG based on `valid' trajectories.}
  \label{fig:gcn}
\end{figure}

\subsection{ObjectNav Module}

\begin{table*}[h]
\caption{Comparison of \textit{ObjectGoal} method with baselines.}
\label{table:comparison_ObjectGoal}
\centering
\begin{tabular}{p{0.3\linewidth}p{0.1\linewidth}p{0.1\linewidth}p{0.1\linewidth}p{0.1\linewidth}p{0.1\linewidth}p{0.1\linewidth}p{0.1\linewidth}}
\hline
Method & Success $\uparrow$ & SPL $\uparrow$ & SoftSPL $\uparrow$ & DTS $\downarrow$\\
\hline
Random & 0.006 &  0.0049 & 0.0363 &  6.6547 \\
Frontier Based Exploration~\cite{yamauchi1997frontier} & 0.598 & 0.3703 & 0.3891 & 4.2478 \\
\textbf{Our Method} & 0.94133 & 0.658 & 0.67931 & 0.3047 \\
\hline
\end{tabular}
\end{table*}
It is imperative that out-of-view object finding is a needed feature of the telepresence system. As an example, suppose a user wants to find where an object (say medicinal supplies) is at a remote location. Instead of manually driving and searching the entire indoor area, this task can help overcome the absence of physical doctors in safe homes in contagion scenarios. A single robot used across multiple patients can carry instruction of different doctors as well as patients. As seen in Fig.~\ref{fig:arch_semnav}, the user gives instruction from the operator (Master) end to find an object. The visual perception (camera feed of scene) of the robot is continuously processed to identify current objects and regions in view. 

A 3-layered Graph Convolution Network (GCN) is trained with spatial relation weights and object finding trajectories that lead to a success. The number of connection layers in GCN (Fig. \ref{fig:gcn}) is selected as three, as more number of edge chains (paths) does not yield enough distinct node level embeddings, as otherwise most nodes will get connected with some other distant node if graph path length is not restricted. The input to the GCN is the spatial relational weights of each object and region as a node along with their adjacency matrix, and output is 128 dimensional node embedding. The relational weights among objects with objects and objects with regions is learned by a combination of Visual Genome~\cite{krishna2017visual} extracted weights and extraction from random agent exploration on large number of AI Habitat scenes. This structure is called Spatial Relational Graph (SRG). The GCN takes two inputs during training: (i) input features for every node i, represented as a $N\times D$ matrix (N: number of nodes, D: number of input features); and (ii) graph structure in the form of an adjacency matrix A of size $N\times N$~\cite{kipf2016semi}. It produces an output of dimension $N\times E$ where $E$ is the dimension of the embedding. The \textit{region} and \textit{object} categorical values are mapped to integer values using the \textit{one-hot encoding vector} to avoid bias, i.e., the index of the node has value $1$ and other values are zeros. At evaluation time, based on a policy of object finding, given visible regions, history of walks and target object, the aforementioned trained model is used to calculate similarity score of current walk with trained walk data of same category of object. During runtime, the actuation move command is sent to the `Robot' for execution using AI Habitat's Geodesic Shortest Path Follower~\cite{shortestPath} navigation planner.  The success rate of this methodology has been found to be around 94\% for realistic indoor scenarios tested for 19 common indoor objects in 6 scene groups of Matterport3D dataset. The results are enlisted in Table ~\ref{table:comparison_ObjectGoal}.

\subsection{AreaNav Module}
\label{secAreaNavModule}
\begin{figure}[t]
    \centering
    \includegraphics[width=\linewidth]{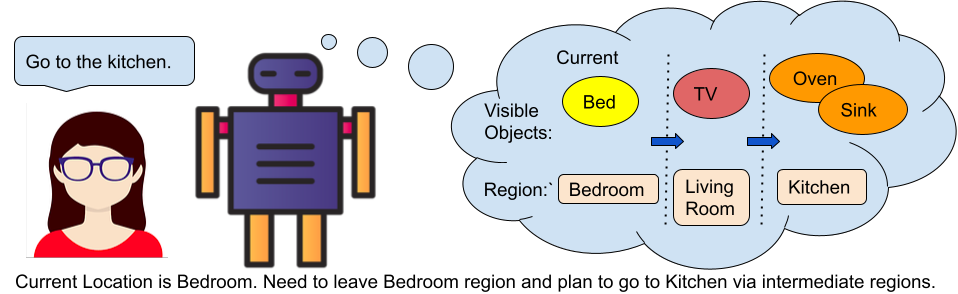}
    \caption{Example of `AreaGoal' task of navigation.}
    \label{fig:areanav_intro}
\end{figure}

The research community has mostly focused on the \textit{ObjectGoal} problem, where as there is no dedicated work for the \textit{AreaGoal} problem. Hence there is no crisp definition of the problem statement and how to decide task completion. Hence, we introduce the task definitions as below. 
\begin{itemize}
\item Definition of Area (A): an area in the context of robot navigation is a region or zone in space where the robot can navigate. Unreachable areas or obstacle-blocked areas are not considered.  
\item Definition of Concrete Boundary (P): The boundary of an area is the region outlining the polygonal sides of it. However, there may be cases like passage, door, openings where there is an overlap between two or more areas. Concrete Boundary marks the line after crossing which there is zero membership of that sub-area to other areas / regions. 
\end{itemize}
Next, we concertize the problem statement and divide it into three problems as follows:
\begin{itemize}
    \item Problem 1: A robot \textit{R} needs to traverse to enter the \textit{concrete} boundary \textit{P} of an area \textit{A}, given a goal task to navigate there. Hence, just getting a view of the area is not sufficient, the robot needs to be within the area for the task to be complete. This we denote as \textit{AreaGoal} task.
    \item Problem 2. The above task completes when the area / region comes into robot perception view. This is a softer \textit{AreaGoal} problem. This can come handy when just the outside view serves the purpose. This we denote as \textit{ViewAreaGoal} task.
    \item Problem 3. The robot needs to navigate to the centroid point or centroid sub-region within a radius of 1.0 m of the goal area's mathematical centroid. However, for this problem, the layout of the area needs to be known beforehand by fusion of external knowledge or learned apriori by exploration. There can be blockage in the center of an area -- meaning no navigable point to go to. In that case the point closest to the navigable centroid is considered. This we denote as \textit{CenterAreaGoal} task. While the earlier problem definitions are for unknown mapless environment, the latter either requires a metric map or run-time map generation based on the approximate navigable center point by taking into view the depth of surrounding structure (like walls) and free space.
\end{itemize}
We present results on the first two problems in a realistic simulation environment of AI Habitat and later test it on real world housing apartments to establish the efficacy of the approach.
\begin{figure}[h]
    \centering
    \includegraphics[width=\linewidth]{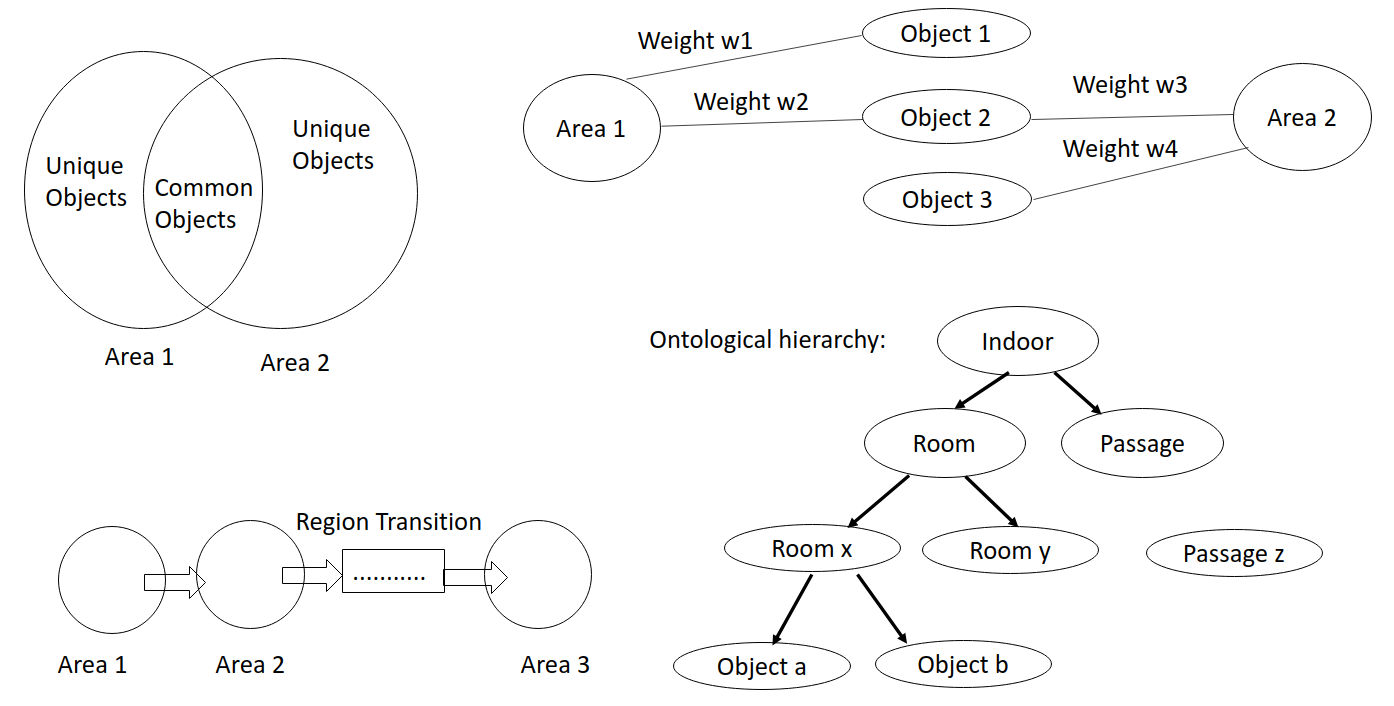}
    \caption{Various ways to represent the concept of area.}
    \label{fig:areanav_reps}
\end{figure}

As shown in Fig. \ref{fig:areanav_reps}, two or more areas can have common objects while some areas will have tight coupled unique objects. The unique objects aid in area identification with high accuracy. Also, region transition paths plays a complementary role in the area search. If a robot is in current location within area A, and target area is T, then based on whether A and T are adjacent or not, or what intermediate areas need to be traversed through -- a navigation planning decision can be taken. Another aspect is related to edge weights between objects and regions as nodes. This spatial relational graph (SRG) will aid in statistically deciding the current area among the set of regions. Finally, a good way to represent indoor environment is by means of ontology, where regions can be divided into passages and room enclosures; and rooms can be subdivided into specific areas like bedroom, toilet. Each area can be represented as a composition of objects. In this paper, to solve the `AreaGoal' task, we have leveraged this concepts.

\subsubsection{AreaNav Methodology}
The main task of `AreaGoal' class of problems can be broken into two subtasks: identifying the area; and navigation from one area to another. Once the robot starts in a random location, first, it needs to identify the area it is currently in. Next, it needs to go out of the current area if it is not the target area. If there are multiple openings from current area, it needs to select the most statistically close one to the target area, and go there. If after taking that choice of path, the area is not reached, it needs to backtrack to an earlier viable branch position to continue the area search. 

\begin{figure}[h]
    \centering
    \includegraphics[width=\linewidth]{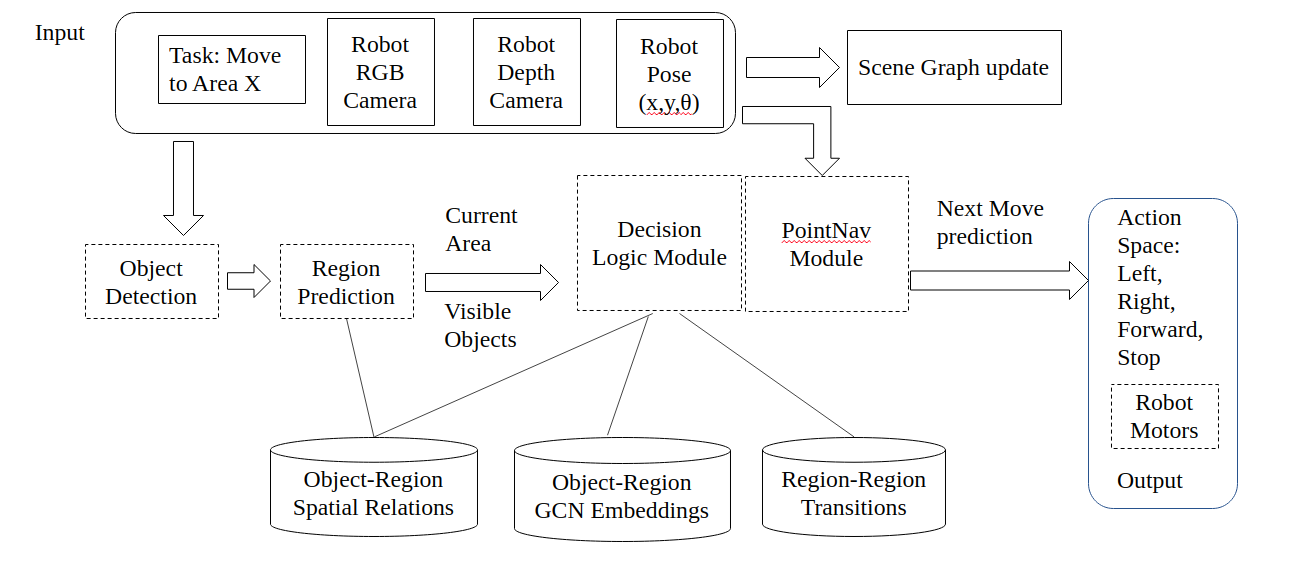}
    \caption{System architecture of AreaNav module comprising submodules.}
    \label{fig:areanav_arch}
\end{figure}

\begin{figure}[h]
    \centering
    \includegraphics[width=\linewidth]{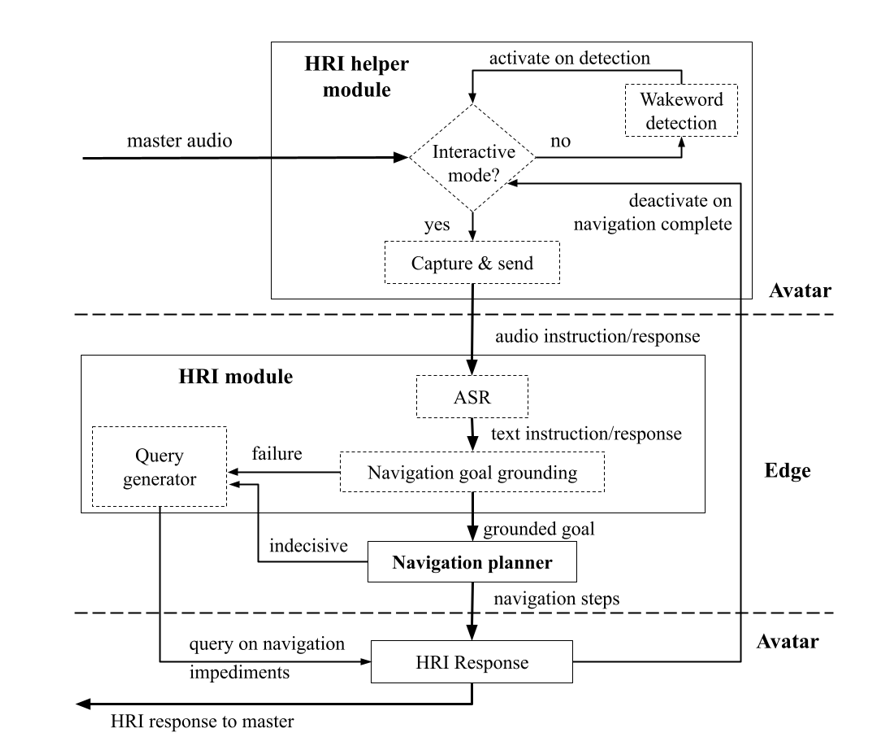}
    \caption{Invocation of HRI module for task understanding and disambiguation.}
    \label{fig:areanav_hri}
\end{figure}

\begin{figure}[h]
    \centering
    \includegraphics[width=\linewidth]{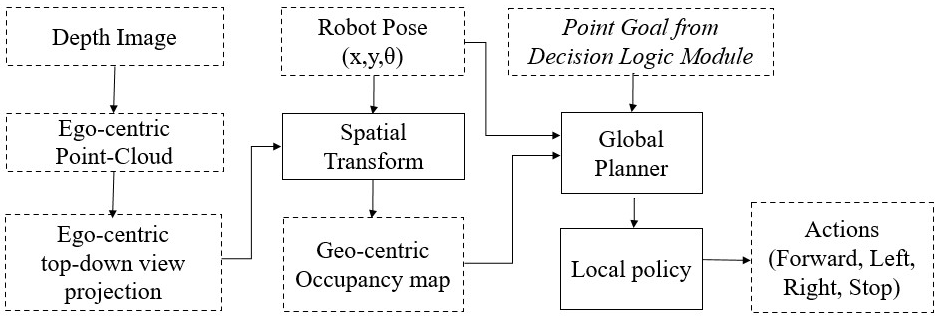}
    \caption{Invocation of `PointGoal' module for navigation towards objects or grid map corners.}
    \label{fig:areanav_pointgoal}
\end{figure}

As shown in Fig. \ref{fig:areanav_arch}, the proposed system comprises of an input space comprising of sensor observations of RGB-D image and odometry readings, while the output is a member of the action space (left, right, forward) with goal of moving to the target area. The target area is specified in human instruction, processed by HRI module as shown in Fig. \ref{fig:areanav_hri}. The HRI module will help in identifying the task type as well as handling cases where robot is stuck at a place and needs human intervention to move towards the target. Based on object sets detected over a stream of scenes, the robot predicts the region based on object-region relations, region-region transitions and learnt GCN embeddings. The GCN was trained using random walks over large number of AI Habitat scenes to extract embedding representation for each node (object and regions). Then, based on aforementioned inputs a decision is taken to move towards target. As navigating from one area to another is a long term goal, it is broken into local goals that are handled by the `PointGoal' module (Fig. \ref{fig:areanav_pointgoal}), discussed in Section \ref{PointNav}. Based on the robot's current pose and the given goal location, the `PointGoal' module plans the shortest navigable path.  Also, when the robot is stuck in an area, an external fair point is given as a goal to the `PointGoal' module for it to explore out of that stuck area. An external fair point is an unvisited point lying within the initialized grid, but at some pre-specified distance away from current robot location. This fair point is sampled from points close to the least sampled corner (based on earlier choices) among the four corners of the grid.


The above task is dependent on several software entities. They are enlisted below.

(a) Region Relation Graph: An indoor space is comprised of objects and areas (regions). There are some specific objects like cup, chair; and there are some generic objects like wall, ceiling, floor which are continuous. There are three types of generic spatial relations: (a) object near another object, like table near chair (b) objects situated within a region, like bed as an object in region `bedroom' (c) regions closely connected with other regions, like regions `dining room' and `kitchen'. The list of indoor objects (as a subset of MS COCO~\cite{lin2014microsoft}) considered are -- bench, bottle, wine glass, cup, fork, knife, spoon, bowl, banana, apple, sandwich, orange, broccoli, carrot, hot dog, pizza, donut, cake, chair, sofa, potted plant, bed, dining table, toilet, TV monitor, laptop, mouse,  remote (TV), keyboard, cell phone (mobile phone), microwave, oven, toaster, sink, refrigerator, book, clock, vase, scissors, teddy bear, hair drier, and toothbrush. The regions (areas or zones) considered are: bathroom, bedroom, dining room, study room, kitchen, living room, toilet, balcony and passage.

Two separate approaches were used to create a weighted relation graph. An entry in that relation graph is `bed' and `bedroom' having very high relationship weight close to $1.0$. In the first approach, these relations were extracted and normalized through a user survey comprising questions and responses such as how close two objects and regions are on an interval scale of 0 to 1. In approach B, the weights were learnt via observations registered in random walks in AI Habitat environment on a large number of indoor realistic scenes. Through ablation studies in various indoor layouts and scenes, it was found that manual survey based relation weights, provided better results for the `AreaGoal' task.

(b) Region Transition Graph:
The `AreaGoal' problem specifically deals with navigating from region `A' to target region `Z' via intermediate regions as per layout of the indoor environments. In this regard, the past history of traversal through regions can guide whether a robot is moving towards a target region. As an example, when navigating from `kitchen' to `bedroom' the robot will have to generally navigate via intermediate region of `dining room'.

(c) Object Category Recognition: Identification of an area is generally determined by the type of objects the area contains. As an example, an area `bedroom' will contain the object `bed' and optionally `chair', `clock', `cell phone', etc. In this regard, MaskRCNN~\cite{bharati2020deep} and YOLO~\cite{jiao2019survey} based approaches trained on MS Coco dataset has been used extensively in literature. However, contrary to simulation environments like AI Habitat, where object category identification can have ground-truth known via semantic annotation, the current framework was been made in a way to work in real world, without using any ground-truth. In our case, robot observations (RGB-D image frames and odometer readings) are taken as input and action is given as output - this is the only communication between the simulation environment (treated as black box) and our modules. Through studies in both black box simulation settings and real world indoor environments, YOLO was found to perform better than Mask-RCNN for real-time robot observation processing in indoors, and hence was adopted here.

\begin{figure}[h]
    \centering
    \includegraphics[width=\linewidth]{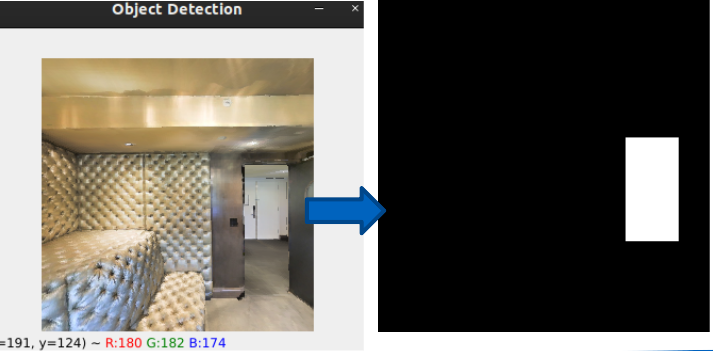}
    \caption{Detection of opening based on rectangular contours by processing depth observation.}
    \label{fig:depth_opening}
\end{figure}

In order for the robot to navigate out of an area, it is important to detect openings. The object detection algorithm (YOLO) was pre-trained for additional 3 object classes: `open door', `closed door' and `free area' (passage). This classes do not exist in off-the-shelf YOLO models. The training was done by manually annotating 100 images per class taken from simulation environment and real world (Google Images) using the Visual Object Tagging Tool (MS VoTT\footnote{https://github.com/microsoft/VoTT}). Another alternative way that was employed was looking for rectangular contours in depth image as shown in Fig. \ref{fig:depth_opening}.

(d) Obstacle Avoidance: Although many robots now a days has inbuilt obstacle avoidance mechanism based on depth sensor, however, it works in a passive way, i.e. when a `move forward' instruction is given and there is an obstacle in front, the robot will try to move, but its sensors will prohibit it from executing the task. Hence, alternatively, a better way is to analyse depth images and post diving the image into `free' and `occupied spaces' based on gray-scale threshold value, the robot is prevented from issuing non-navigable commands, thereby saving step counts.

(e) Reflective Surface Avoidance: A typical indoor house may consist of a large number of mirrors and reflective surfaces. If a robot relies on its depth sensor for finding free spaces, a mirror will give wrong depth information resulting in robot collision. Hence, two complementary strategies are taken. Firstly, a check is kept if current and past RGB image observations across robot steps are very similar using a threshold parameter. Secondly, the approach of Mirror3D \cite{tan2021mirror3d} is adopted to identify reflective surfaces and correct depth estimates to aid robot in successful navigation.

(f) Object to PointGoal Mapper: When an object say `bed' related to target region `bedroom' comes in view of the robot, it needs to move towards `bed' to maximize its chance of being within the target region. This is done by a mapper module, that takes the RGB-D bounding box of identified object and maps it to a point on the 2D navigable surface. This point can be passed to a `PointGoal' local planner (like ROS Nav2 Planner 
to plan the route avoiding obstacles.

(g) PointGoal Planner: As discussed in earlier subsection \ref{PointNav}, the point to point navigation will be done by a `PointGoal' planner. This module also aids in planning a path to a distant grid point, when the need is to go out of a region, and no openings are detected by aforementioned object recognizer methods.

(h) Visible Area Identifier: In the `ViewAreaGoal' task, the task is said to be complete when the target area comes in view or objects related to that area comes in view. This is accomplished by seeing related objects to target region in subsequent image frames with high confidence, provided the objects are at a distance from current robot position. Sometimes, it may happen that due to occlusion or failure in object detection algorithm, the robot needs to enter the area in order for related objects to be detected. Then this problem reduces to standard `AreaGoal' task. 

(i) Area Identifier: In the `AreaGoal' task, the robot does a full $360^0$ rotation at intervals. The rotation is triggered either by (a) rotation interval counter or (b) when it is detected that robot has entered a new region (based on passing detected openings) or (c) objects dissimilar to last region is detected. The robot invokes the edge values of the Region Relation Graph for each of the objects O in the rotation view to predict the most probable region R as shown below:
\begin{equation}
R = max (\sum ( p(O_x) * p(R_y) * c(O_i) * p(R_y \mid O_x) / N) )    
\end{equation}
Here, p($O_x$) denotes probability of an object x based on frequency of occurrence among all objects. Similarly p($R_y$) denotes the probability of a region among all regions based on the frequency of occurrence in observations. c($O_i$) denotes confidence score of the object detector for that object class. $p(R_y \mid O_x)$ denotes conditional probability that robot is in one of `y' regions based on the view of a single object `x'. While doing a $360^0$ rotation, if an object with high affinity to a region comes in view with good detection confidence score, the rotation is stopped and area inferred. This saves step counts and unnecessary full rotations.

(j) Area Center Identifier: Where as the `AreaGoal' task is completed when the robot enters the target area, in case of `AreaCenterGoal' task, the robot needs to be close to the center of target area or its closest navigable point. Once target area is identified using aforementioned methodology (i), next the robot needs to navigate to a center point. This can be achieved by invoking the mapper module to identify the coarse boundary from depth observation and estimating a center point that is navigable. A simple implementation will be a heuristic based estimation to calculate a point and pass it on to `PointNav' module. The point in 2D plane can be calculated as follows:\\
C = ( centroid ( visible objects' average depth points) + centroid (coarse region map perimeters) ) / 2\\
In future work, we will tackle this problem separately as it involves advances in relative pose estimation and depth sensor noise correction. We have kept this task out of scope of this paper.

(k) Backtracking to Landmark: The Visual Semantic Map as discussed in Section \ref{visSemMapsection}, helps in keeping track of unexplored places in a grid. It also marks certain places as landmarks having more than one branches (openings) and objects having high affinity with target region. In case, the robot meets a dead end or has followed a branch path, instead of exploring, it can directly invoke the `PointNav' module to go to the nearest unexplored landmark grid cell.

The high-level algorithmic workflow for the Area Goal task is shown in Algorithm 1. Initially a square grid map with empty cells is initialized with 4 corner points. At first, the robot performs a rotation to identify current area, and do area prediction at intervals or when robot observes objects closely tied with target area. If there are mirrors or blockages, the robot turns around for new views. Once robot identifies that current area is not target area, it searches for openings and free space to enter newer areas. Intermittently, it moves towards objects having highest relation to target area. In case of dead end, robot backtracks to last stored landmarks (unexplored openings or unexplored object directions in the map). Finally, if landmarks are exhausted, robot tries to plan path towards the four corners of the grid map one by one, expecting that target area will lie in one of the paths. It is to be noted that grid map gets updated with new information at each observation.

\begin{algorithm}
\label{algoArea}
\caption{Pseudocode of AreaGoal task}\label{algoAreaGoal}
\begin{algorithmic}[1]
\State \textbf{Parameters}:\\ 
img $\leftarrow$ RGB-D camera egocentric image stream;\\
actuation $\leftarrow$ commands to robot wheels;\\
sf $\leftarrow$ link to various software modules;\\
ta $\leftarrow$ target area as per instruction\\
map $\leftarrow$ generated map for storing landmarks\\
odom $\leftarrow$ location of the robot in 2D space\\
\textbf{Initialization}:\\
map $\leftarrow$ empty N*N size grid map, say N = 30m\\
corners $\leftarrow$ four corner co-ordinates of grid map\\
steps $\leftarrow$ 0 // step count of the actuation moves
rotateCount $\leftarrow$ 0; rotateFlag $\leftarrow$ 1; taflag $\leftarrow$ 0
\While{ta NOT reached AND steps $\lt$ 500 }
    \State Wait for actuation completion;
    \If{rotateFlag == 1 AND rotateCount == 0}
        \State do a $360^0$ rotation to scan area
        \State area $\leftarrow$ sf.areaPredict(img sequences)
        \If{area == ta}{ taflag $\leftarrow$ 1; break;}
        \EndIf
    \EndIf    
    \If{sf.estimate(img) finds mirrors / block}
        \State do a 90$^0$ rotation towards \{left or right\}
    \EndIf      
    \If{sf.estimate(img) finds openings X}
        \State sf.PointGoal(X$_1$ bounding box's center)
    \EndIf
    \If{software.YOLO(img) contains objects}
        \If{probability(area $\mid$ object) $\gt$ 0.9}
            \State rotateFlag $\leftarrow$ 1; rotateCount $\leftarrow$ 0;
            \State return (to do rotation);
        \Else{ sf.PointGoal(object most related to ta)}    
        \EndIf
    \Else
        \If{map.landmark is NOT exhausted}
            \State sf.PointGoal(nearest landmark)
        \Else    
            \State sf.PointGoal(corners[least accessed])
        \EndIf
    \EndIf
    \If{rotateCount$\gt$ 20}{ rotateCount $\leftarrow$ 0}
    \EndIf
    \State update map(odom, img); rotateCount += 1;
\EndWhile
\If {taflag == 1}
    \State print `Area \$ta reached in \$steps steps';
\Else
    \State print `Task Incomplete after 500 steps';
\EndIf    
\end{algorithmic}
\end{algorithm}

\subsubsection{AreaNav Experimental Results}

20 scene groups of Matterport 3D (MP3D) dataset were used to devise tasks for different target regions (areas). Ground truth information of the regions were used as boundaries of areas. Visual inspection was also carried out in scene groups giving poor results to identify the underlying cause, and adapt the algorithm heuristics. Through ablation studies, it was found contrary to `ObjectGoal' task, learnt GCN embeddings do not enhance `AreaGoal' task - hence it is not used in the baseline experiments. Considering MP3D house layouts being quite large, the upper limit of step count was kept at 500, by which, if the robot is unable to reach the target area, the search terminates as failure. 

As can be seen from the results Table~\ref{table:comparison_with_baselines}, we take note of both success and step count. Success as a metric is important to tell whether a random movement can ever reach a target area in allowed steps. Next metric step count tells how quickly the specified method based navigation reached the target area. Hence, for random movement based navigation we also get success many times, but the step count is large compared to other methods. This gives us an insight, that there are certain areas which are difficult to navigate to (if at all can be navigable autonomously for a specified layout), whereas there are certain areas that are simpler to traverse to.

\begin{figure*}[t]
    \centering
    \includegraphics[width=\linewidth]{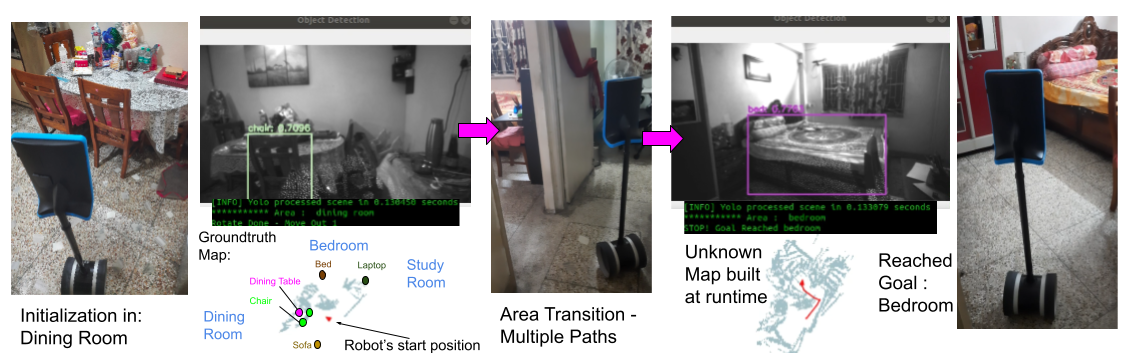}
    \caption{Real life example of `AreaGoal' task. Starting at dining room, the robot needs to navigate to bedroom.}
    \label{fig:areanav_real}
\end{figure*}

The proposed method was also successfully tested in real life settings of indoor home. For use in real world with noise, depth sensor distance up to 2.5 meters were considered for map building or object localization. Fig. \ref{fig:areanav_real} shows an example snapshot of the robot navigating to the `bedroom' area after getting initialized at `dining room' region.
\begin{figure}[h]
    \centering
    \includegraphics[width=\linewidth]{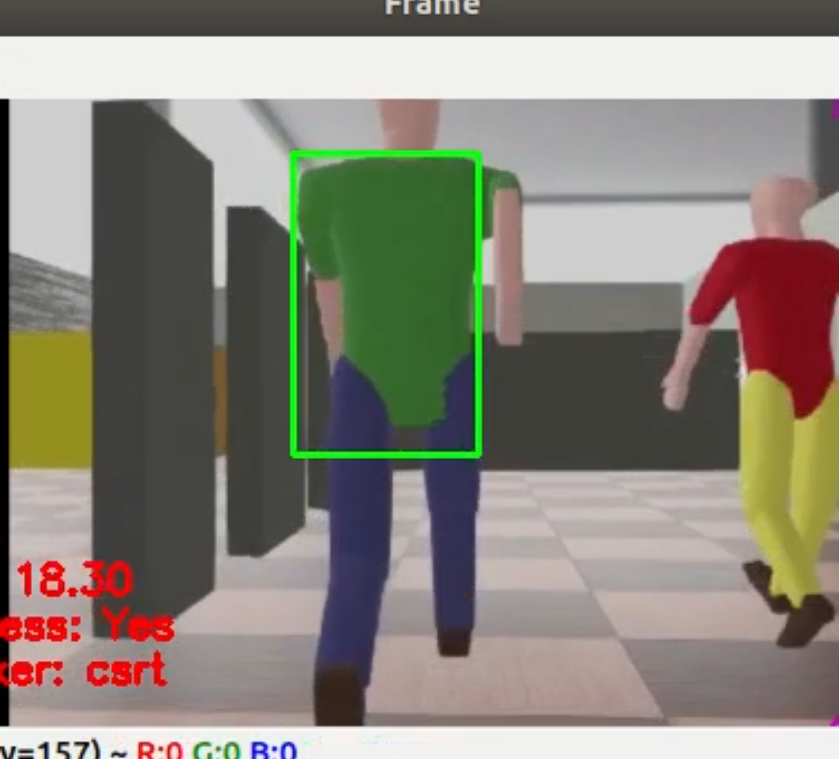}
    \caption{Person following in Webots with 2 persons.}
    \label{fig:personFollowWebots}
\end{figure}
\begin{figure}[h]
    \centering
    \includegraphics[width=\linewidth]{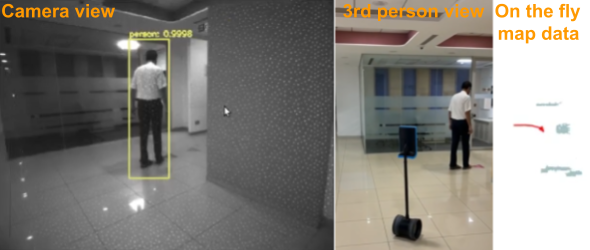}
    \caption{Person following in real world office settings.}
    \label{fig:personFollowReal}
\end{figure}

\begin{table*}[t]
\caption{Comparison of above \textit{AreaGoal} method with baselines for different target areas (Goal).}
\label{table:comparison_with_baselines}
\centering
\begin{tabular}{p{0.1\linewidth}p{0.2\linewidth}p{0.2\linewidth}p{0.1\linewidth}p{0.15\linewidth}p{0.07\linewidth}}
\hline
Target & Method & AreaViewGoal Task: & & AreaGoal Task:\\
goal &  & Success $\uparrow$ & Step $\downarrow$ & Success $\uparrow$ & Step $\downarrow$\\
\hline
Bathroom & Random & 0.9 &  230 & 0.9 &  212 \\
or & Frontier-based~\cite{yamauchi1997frontier} & 0.9 & 128 & 0.9 & 147 \\
Toilet & \textbf{Our Method} & 0.95 & 110 & 0.95 & 122 \\
\hline
Bedroom & Random & 0.85 & 354 & 0.8 & 423 \\
 & Frontier-based & 0.95 & 178 & 0.95 & 182 \\
 & \textbf{Our Method} & 0.95 & 125 & 0.95 & 136 \\
\hline
Dining & Random & 0.9 & 290 & 0.9 & 244 \\
room & Frontier-based & 0.95 & 240 & 0.95 & 246 \\

 & \textbf{Our Method} & 1.0 & 204 & 1.0 & 220 \\
\hline
Study & Random & 0.5 &  442 & 0.3 & 489 \\
room & Frontier-based & 0.7 & 390 & 0.65 & 430 \\
 & \textbf{Our Method} & 0.9 & 280 & 0.85 & 343 \\
\hline
Kitchen & Random & 0.9 &  290 & 0.9 & 301 \\
 & Frontier-based & 0.95 & 157 & 0.95 & 173 \\
 & \textbf{Our Method} & 1.0 & 122 & 1.0 & 147 \\
\hline
Living  & Random & 0.6 & 482 & 0.55 & 497 \\
room & Frontier-based & 0.85 & 137 & 0.85 & 143 \\

 & \textbf{Our Method} & 0.95 & 110 & 0.95 & 119 \\
\hline
Average & Random & 0.775 &  332 & 0.725 & 361 \\
across & Frontier-based & 0.83 & 205 & 0.875 & 221 \\
areas & \textbf{Our Method} & 0.958 & 159 & 0.942 & 182 \\
\hline
\end{tabular}
\end{table*}

\subsection{Person Following}
\label{secPersonFollow}
\begin{algorithm}
\caption{Pseudocode of Person Following Robot}
\label{algoPersonFollow}
\begin{algorithmic}[1]
\State \textbf{Parameters}:\\ 
image $\leftarrow$ RGB camera egocentric image stream;\\
actuation $\leftarrow$ commands to robot wheels;\\
software $\leftarrow$ link to software modules;\\
target $\leftarrow$ target person bounding box to follow;\\
sa $\leftarrow$ area of scene (eg. Left, Middle, Right);\\
\textbf{Initialization}:\\
persons $\leftarrow$ bounding boxes from software.YOLO(image);\\
target $\leftarrow$ select a person's bbox from view;\\
features $\leftarrow$ software.extract(target);\\
tracker1 $\leftarrow$ software.GOTURN(features);\\
tracker2  $\leftarrow$ software.CSRT(features);\\
model $\leftarrow$ software.OneClassSVM(features);\\
lastMove $\leftarrow$ Stop; weight $\leftarrow$ 1/3;
\While{No command to stop Person Follow task}
\State Wait for actuation completion;
        \If{software.YOLO(image) contains persons}
            \State a $\leftarrow$ software.detect(persons, tracker1); 
            \State b $\leftarrow$ software.detect(persons, tracker2);
            \State c $\leftarrow$ software.detect(persons, model);
            \State target $\leftarrow$ voting( a/3, b/3, weight * c );
            \State weight $\leftarrow$ 1/3;
            \State bbox  $\leftarrow$ bounding box dimension of target
            \If{bbox $\geq$ threshold t}
                \State actuation $\leftarrow$ STOP;
            \Else
                \State m $\leftarrow$ max. overlap of target with sa
                \State actuation $\leftarrow$ sa[m];
                \State lastMove $\leftarrow$ actuation;
                \State model $\leftarrow$ software.update(target);
                \EndIf
        \Else
            \State actuation $\leftarrow$ lastMove; weight $\leftarrow$ 1;
        \EndIf
\EndWhile
\end{algorithmic}
\end{algorithm}
One of the important features of an embodied AI based Telepresence robot is to follow a target person when instructed to do so. This feature becomes useful when an elderly person needs to be monitored while doing some activity. Another use case is when a caregiver needs to be followed by a robot between different wards, the doctor being remotely connected over internet. Fig. \ref{fig:personFollowWebots} shows how a person is followed based on first person ego view in Webots, a ROS compliant simulation environment. The choice of Webots was made as a test ground, as AI Habitat 2 did not allowed addition of person objects with physics based trajectories. Fig. \ref{fig:personFollowReal} depicts person following algorithm working in real world settings. The algorithm \ref{algoPersonFollow} for person following is enlisted below. Initially the scene is processed using YOLO object detector to detect persons; and the user is asked to tag a person who needs to be followed. The features of the person selected (bounding box) is passed on to (a) Discriminative Correlation Filter Tracker (CSRT)~\cite{farkhodov2020object} algorithm (b) deep learning based GOTURN tracker~\cite{held2016learning} (c) one class Support Vector Machine (SVM)~\cite{cyganek2009framework}, trained by data augmentation techniques of the single snapshot of the user tagged person pixels of bounding box. The training continues in online fashion while the tracker identifies the person in subsequent frames. The training stops when tracker confidence is low or person goes out-of-view. The former two trackers help in tracking a person object from a stream of sequential image frames. However, when the person gets out of view due to speed, or occlusion and reappears, this SVM based person re-identification method helps in detecting the specific person, where the former two trackers fail. The final target bounding box is decided based on a weighted voting amongst the tracker and the SVM model. In case of person becomes out of view, the SVM model weight matter more as shown in the algorithm. Generally, for such scenarios the robot follows the last trajectory move. Finally, if even then person is not found, then a $360^0$ rotation followed by random moves is the way out. At each frame, the scene area is divided into three parts (Left, Forward, Right) and based on maximum overlap of bounding box of target person with scene area, the decision to move in which direction is taken. To avoid collision with target person, and maintaining a safe distance while following, the bounding box dimensions of target person is tracked to see if it lies within a threshold. If the threshold is crossed, the robot stops till the person moves away from close proximity. Simultaneously, map data is also learned to avoid static obstacles in future traversals.
\section{User Interface and User Study}
\label{secUI}
\begin{figure*}[h]
\centering
\includegraphics[width=\linewidth]{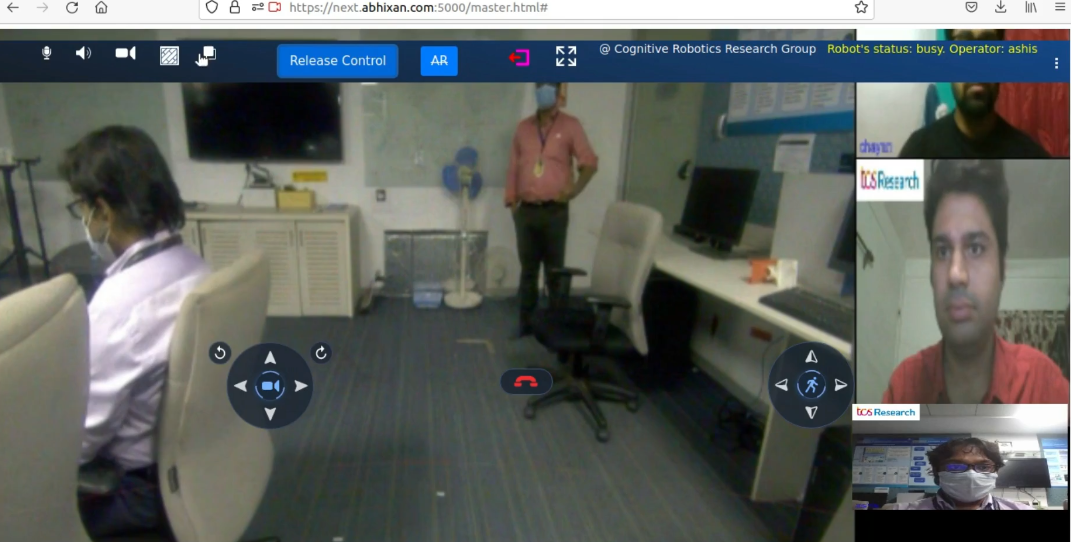}
\caption{Browser based user interface for remote user.}
\label{fig:TeledriveUI}
\end{figure*}
To support ease of use and portability, a browser based user interface (as shown in Fig. \ref{fig:TeledriveUI}) is developed. The user joins over the internet using a secured authentication process. The background video is a teleconference setup with the main video stream being of the robot's front camera.

The controls, visible as icons in the figure, from left to right are: (1) Microphone - to control the audio of remote user, (2) Speaker - to control the hardware speaker of device on which browser interface is loaded, (3) Video - to turn on or off the self camera view, (4) Map - to display pre-created map for this location (if any). This in turn will help in map based point goal navigation. (5) Map Size control - to zoom in and zoom out from the map (6) A toggle button named `Take Control' or `Release Control' - for the remote user to take control of the robot located in a different location. In the screenshot, user named `Ashis' can be seen in control of the robot. (7) `AR' toggle button to enable a user to click on the screen to create an augmented reality icon marker, where the robot should move to (8) Exit icon - to sign off from current session (9) Fullscreen - to hide controls and get full view of robot's camera feed. The central 3 controls are as follows: (1) Camera tilt icon set - to tilt or rotate the camera (2) Hangup button - to end the session of the teleconferencing call (3) Robot control - to move the robot in all four directions, with center being the `Stop' command. Similar to existing real-time conferencing software, the right hand side is dedicated to display other remote users logged into the session. 
\begin{figure*}[h]
\centering
\includegraphics[width=\linewidth]{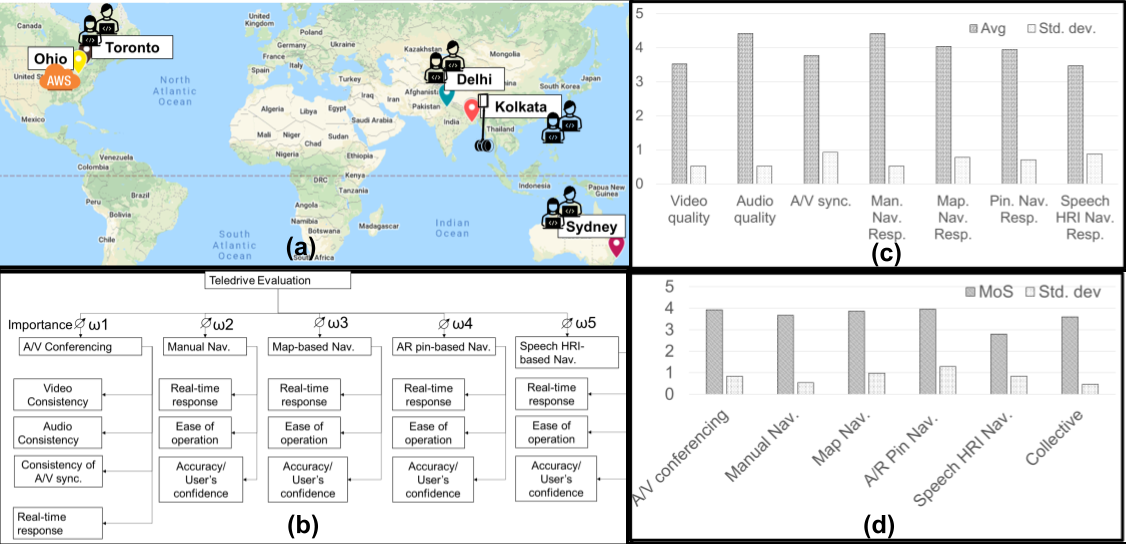}
\caption{(a) Varied remote location of users (b) Survey questionnaire on which ratings were given (c) Response time feedback for the surveyed features (d) Quality of user experience for the surveyed features.}
\label{fig:userStudy}
\end{figure*}

On a study conducted over 30 users across different demographics and geographic locations, the system was found to be responsive and easy to use. The study results are listed in Fig. \ref{fig:userStudy}. The 30 remote users were located at different cities across the world and they were asked to login from their device browsers to test the features of the Teledrive solution and rate their experience in survey questionnaire on a scale of 1 to 5 (5 being excellent). The response time for each feature was calculated based on a user clicking a button or giving a command on the remote user side and the corresponding trigger in the robot and is listed in the above figure -- this was calculated without user intervention. The user study tested the following features: (a) Audio-Video Conferencing over WebRTC (b) Manual Navigation (c) Map based Navigation (d) Pin Goal Navigation (dropping a pin) (e) Speech based HRI to move to some location like object for ObjectGoal and area for AreaGoal. The user was a given a option to select the relative importance of a feature. The feature-wise collective Mean Opinion Score (MOS) computed as:  
\begin{equation}
    M_k =  \frac{ \sum_{j=1}^{N} \frac{w_{kj} \sum_{i=1}^{A_k}{a_{ij}}}{A_k} }{\sum_{j=1}^{N}w_{kj}}
\end{equation}
Here, $M_k$ = MOS for the kth feature; $a_i$ = rating of the i-th attribute of the k-th feature by j-th user; $A_k$ = total number of attributes in k-th feature; $w_{kj}$ = importance of k-th feature for the j-th user; N = total number of users.

The robot side interface is similar to Master (remote) side except lack of controls which are unnecessary. The robot side interface is needed in order for the robot side user (say a patient or elderly person) to see the remote users (say a team of doctors) to perform some basic operations like audio-video control on the robot side. As a security feature, the Master operator has no direct control on the edge server where ROS 2 is running with software modules as ROS nodes. It is through the frontend communication channeling that Master can invoke the software feature executions.

\section{Conclusion}
\label{secConclusion}
In this article, we presented an architecture for embodied AI tasks targeted toward a telepresence setup, especially in patient care scenarios. We have described approaches of individual embodied AI modules with a specific focus on the `AreaGoal' and `person following' problem. We also back our claims with relevant results. In future work, we expect to adjoin additional embodied AI modules for the tasks of language-grounded navigation, contextual scene understanding, gesture-based navigation, etc. Apart from presented use cases, this work will be expanded for deployment in office conference scenarios, retail stores, and factory workshops requiring supervision tasks. 

\bibliographystyle{bst/sn-standardnature}
\bibliography{sn-bibliography}

\section{Statements and Declarations}

\subsection{Acknowledgements}
Thanks to Dr. Balamuralidhar Purushothaman of TCS Research for guidance and motivation for this work.

\subsection{Funding}
The authors have received research support from Tata Consultancy Services Limited, India.

\subsection{Code and Data Availability}
Code is IP protected, and data on which experiments are done is available for download from the link: \url{https://github.com/facebookresearch/habitat-lab\#data}.

\subsection{Author Contributions}
Conceptualization: [Snehasis Banerjee, Abhijan Bhattacharya]; Methodology: [Snehasis Banerjee, Sayan Paul, Ashis Sau, Pradip Pramanick]; Writing and Content: [Snehasis Banerjee, Abhijan Bhattacharya, Ruddradev Roychowdhury, Chayan Sarkar]; Supervision: Brojeshwar Bhowmick.

\subsection{Ethics Approval}
Not Applicable

\subsection{Consent to Participate}
Informed consent was obtained from all individual participants included in the user study.

\subsection{Consent to Publish}
User study data is anonymized and not published in this work.

\section{Appendix}
\label{appendix1}
The embodied navigation metrics are presented below.

1. Success : It is the ratio of the successful episode to total number of episodes. An episode is successful if the agent is at a distance $\leq$ 1.0 m from the target object at the end of episode.

2. SPL (Success weighted by path length) : It measures the efficiency of path taken by agent as compared with the optimal path. This is computed as follows:
\begin{equation}
SPL=\frac{1}{N} \sum_{i=1}^{N} S_i . \frac{l_i}{max(p_i, l_i)}
\end{equation}
\noindent
where N is the number of test episodes, S$_i$ is a binary indicator of success, l$_i$ is the length of the shortest path to the closest instance of the goal from the agent’s starting position and p$_i$ is the length of the actual path traversed by the agent. SPL ranges from 0 to 1. Higher SPL indicates better model performance for the following reasons,
\begin{itemize}
\item High SPL signifies that the agent trajectory length is
comparable to the shortest path from the agent’s starting
position to the goal.
\item SPL also suggests that the agent has reached the
closest instance of the target goal category `t' from its starting
position.
\end{itemize}

3. SoftSPL: One of the shortcomings of SPL is that it treats all failure episodes equally. This is addressed in the SoftSPL metric by replacing $S_i$, the binary indicator of success (whether goal reached or not as 0 or 1), by \textit{episode\_progress}, a continuous indicator. This \textit{episode\_progress} ranges from 0 to 1 depending on how close the agent was to the goal object at episode termination.
\begin{equation}
SoftSPL=\frac{1}{N} \sum_{i=1}^{N} ( \underbrace{1 - \frac{d_i}{max(l_i,d_i)}}_{episode\_progress}) . (\frac{l_i}{max(p_i,l_i)})
\end{equation}
\noindent where N is the number of test episodes, $l_i$ is the length of the shortest path to the closest instance of the goal from the agent’s starting position, $p_i$ is the length of the actual path traversed by agent and $d_i$ is the length of the shortest path to the goal from the agent’s position at episode termination. Similar to SPL, higher SPL indicates better model performance.

4. Distance to Success (DTS) : It signifies the distance between the agent and the permissible distance to target for success at the end of a search episode.
\begin{equation}
DTS = max( (\left \| x_T-G \right \|_2 - d) , 0 )
\end{equation}
\noindent where $x_T$ is the $L2$ distance of the agent from the Goal at the end of the episode, $d$ is the success threshold. A lower DTS value at episode termination indicates that the agent is closer to the goal category. Therefore, a lower DTS value indicates better performance.

5. Step count: In the AI Habitat environment, the robot action space consists of three actions - turn right by 30 degrees, turn left by 30 degrees, and move forward by 0.25 metres. Each such action is counted as one step. A single rotation is also considered as a step. A full $360^0$ rotation equates to 12 total steps. The less the number of steps taken, the better is the method. Hence, step count is considered as a metric. 
\end{document}